\documentclass[letterpaper,10pt,conference]{ieeeconf}  

\IEEEoverridecommandlockouts                              

\usepackage{booktabs}
\usepackage{url}
\usepackage{amsmath,amssymb}
\usepackage{amsfonts}
\usepackage{bbold}
\usepackage{xcolor}
\usepackage[english]{babel}
\usepackage{graphicx}
\usepackage{subcaption}
\usepackage{algorithm}
\usepackage{comment}
\usepackage{float}
\usepackage{hyperref}
\usepackage[noend]{algpseudocode}

\newcommand{\rev}[1]{{#1}}

\renewcommand{\baselinestretch}{0.974}

\title{\LARGE \bf
Flocking-Segregative Swarming Behaviors using Gibbs Random Fields
}

\author{Paulo Rezeck$^{1}$ and Renato M. Assunção$^{1}$ and Luiz Chaimowicz$^{1}$
\thanks{Paulo Rezeck, Renato M. Assunção and Luiz Chaimowicz are with the Department of Computer Science, Universidade Federal de Minas Gerais,
        Brazil
        {\tt\small \{rezeck,assuncao,chaimo\}@dcc.ufmg.br}.
        This work was partially supported by CAPES, CNPq, and Fapemig.}%
}

\begin{document}

\maketitle
\thispagestyle{empty}
\pagestyle{empty}


\begin{abstract}  
This paper presents a novel approach that allows a swarm of heterogeneous robots to produce simultaneously segregative and flocking behaviors using only local sensing. These behaviors have been widely studied in swarm robotics and their combination allows the execution of several complex tasks.
Our approach consists of modeling the swarm as a Gibbs Random Field (GRF) and using appropriate potential functions to reach segregation, cohesion and consensus on the velocity of the swarm. Simulations and proof-of-concept experiments using real robots  are presented to evaluate the performance of our methodology in comparison to some of the state-of-the-art works that tackle segregative behaviors.
	
\end{abstract}

\section{Introduction}
\label{sec:intro}

Due to the advances in technology that have enabled the mass production of increasingly smaller robots~\cite{miskin2020electronically}, control methods that yield desired collective behaviors using simple local interactions have received much interest in recent years. Inspired by the emergent behaviors commonly observed in nature, one of the main goals of swarm robotics is to develop such methods in a decentralized and scalable fashion, mainly relying on local sensing and communication capabilities.

In this sense, one of the most fundamental mechanisms a robot swarm must exhibit is the ability of group formation and cohesive navigation~\cite{brambilla2013swarm}. Segregation is a particular type of group formation in which robots with common characteristics are placed together and set apart from other groups~\cite{santos2014segregation}. Several applications can benefit from using these behaviors, such as area coverage, surveillance and reconnaissance, transport, foraging, among others.


This work presents a novel stochastic and decentralized approach that allows a swarm of heterogeneous robots to achieve simultaneously segregation and flocking behaviors using only local sensing. To the best of our knowledge, this is the first work to tackle these behaviors together starting from a random initial state and using only local information.
%
%
%
Our approach consists of modeling the robot swarm as a Gibbs Random Field (GRF) defining its potential energy as a combination of Coulomb-Buckingham potential and kinetic energy. Such concepts have been extensively used in statistical mechanics and quantum mechanics to model particle interactions, \rev{ but we revisit them in the swarm robotics context.} 
%
As a consequence of using GRF and such potentials, besides supporting the segregation and navigation of different groups avoiding collision with obstacles, the approach allows the swarm to reach configurations sufficiently close to the global minimum energy. 

Using simulated experiments, we contrast the methodology with a deterministic gradient descent-type algorithm using potential differentials showing that such mechanism is easily trapped at local minima of potential. \rev{In addition, we compare our segregative behavior with some of the state-of-the-art approaches and evaluate the flocking behavior in the presence of noise.}
Real experiments were also performed as \rev{a} proof-of-concept for our GRF approach.

\section{Related Work}
\label{sec:relatedwork}

Most works in swarm robotics usually focus on homogeneous systems, in which all robots have the same characteristics~\cite{dudek1996taxonomy}.  However, in recent years, there is a growing interest \rev{in} heterogeneous systems. As a consequence, new types of swarming behaviors have been investigated, such as segregation. One of the first works to deal with this problem was proposed by Groß et al.~\cite{gross2009segregation}.
\rev{The authors presented a control algorithm inspired} by a collective phenomenon in which segregation occurs by constantly shaking a mixture of different sizes particles (Brazil Nut effect). This study was later extended by Chen et al.~\cite{chen2012segregation} and Joshi et al.~\cite{joshi2019segregation}, when they evaluated and improved performance and presented experiments with real robots.

Another approach to segregate a swarm of heterogeneous robots was presented by Kumar et al.~\cite{kumar2010segregation}. The authors took inspiration from a biological theory that explains how differences in cell adhesion generate mechanical forces that drive cellular segregation (Differential Adhesion Hypothesis)~\cite{steinberg1963reconstruction}. 
This mechanism is modeled with the concept of differential potential, in which robots are subjected to differential artificial potential fields according to their groups.
\rev{The method's convergence} is guaranteed for two classes, but the swarm may be trapped in local minima when more classes are employed. This approach was later extended in Santos et al.~\cite{santos2014segregation, santos2020spatial} to deal with more than two groups of robots.
One limitation is the requirement that robots must have global knowledge about the positions of other robots.

Motivated by the use of differential artificial potential fields, Ferreira Filho and Pimenta~\cite{edson2015segregating} proposed a novel controller that differs from the previous ones by using abstractions~\cite{belta2004abstraction} to represent each group. One advantage of such a controller is that it may not require that all robots receive information from all other robots all the time. More recently, the authors extended this controller to incorporate a collision avoidance scheme \rev{that} does not interfere \rev{with} the original segregation controller~\cite{ferreira2019abstraction}. In a different work~\cite{ferreira2020segregation}, they presented a decentralized control strategy to segregate heterogeneous robot swarms distributed in curves using consensus protocols and heuristics to compute the traveled geodesic distances on curves. This approach assumes that robots know the curve and maintain an underlying fixed communication topology.

Recently, two works assuming minimal and local-only requirements for segregating a swarm of heterogeneous robots have been proposed. Mitrano et al.~\cite{mitrano2019minimalistic} extended the concept of a minimalistic reactive controller~\cite{st2018circle} to achieve segregation. They demonstrate that robots with only a ternary sensor and a controller that maps sensor readings to wheel speeds \rev{can reach} 
a segregated state. Also, considering local sensing and a memory mechanism supported by communication, Inácio et al.~\cite{inacio2019pso} proposed a strategy that combines concepts of Particle Swarm Optimization (PSO)~\cite{eberhart1995PSO} with the Optimal Reciprocal Collision Avoidance (ORCA)~\cite{Berg11} to archive segregation.

Following the works that rely only on local information, we assume that \rev{our} robots only use \rev{their neighbors' relative position and velocities} 
to achieve simultaneous segregation and cohesive navigation. 
\rev{However, most of these works only consider collisions with other robots. Our method also allows the robots to avoid collisions with obstacles in the environment, considering that they are equipped with range sensors, such as infrared sensors.}


\rev{Besides dealing with the segregation problem, our approach also generates a cohesive navigation behavior for the different groups of robots. One of the main challenges in achieving such behavior is reaching a consensus on each part of the group's velocity as its size increases. In addition to that, the groups must remain segregated while navigating.}

One of the earliest and most influential approaches to steer a swarm of homogeneous agents using only local interactions was proposed by Reynolds~\cite{reynolds1987flocks}. This mechanism, called {\em boids}, \rev{combines} 
three simple rules: separation, cohesion, and alignment. 
While most works on this subject deal with homogeneous groups, some \rev{use} 
heterogeneous robotic swarms to study the flocking of distinct groups. Momen et al.~\cite{momen2007mixed} extended the flocking mechanism with heterospecific attraction rules~\cite{monkkonen1990numerical} to model different attraction forces between two \rev{groups} of robots, producing a mixed-species flocking. Ducatelle et al.~\cite{ducatelle2011self} proposed a mechanism that emerges cooperative self-organized behaviors to solve complex tasks using simple local interactions between the robots of the two different groups. Another study on self-organized flocking explored the concept of Swarm heterogeneity in the sense that robots with more capabilities 
help others that lack some capabilities in order to yield the desired behavior~\cite{stranieri2011self}.

Some works tackle the problem of segregated navigation, in which the robots start \rev{in} a segregated state and have to maintain the segregation during navigation. For example, Santos et al.~\cite{santos2014segregative} introduces a novel concept called Virtual Group Velocity Obstacles that combines the concepts of flocking~\cite{reynolds1987flocks} and Velocity Obstacles~\cite{fiorini1998motion} with abstractions to represent the groups. To improve performance over such approach, Inácio et al.~\cite{inacio2018united} proposed the combination of the Optimal Reciprocal Collision Avoidance algorithm~\cite{van2011reciprocal} with the concepts of flocking.

\rev{Different} from these works, our approach simultaneously generates segregation and flocking behaviors. The robots start in a completely random state and, as they move around, they segregate into different groups and keep this segregation while navigating.  To the best of our knowledge, this work is the first to present a fully decentralized stochastic controller that performs both behaviors using only local interactions. 

Moreover, although our approach produces flocking behaviors, we do not use or extend the mechanism proposed by Reynolds~\cite{reynolds1987flocks}. We model the swarm using dynamic Gibbs Random Fields (GRF), which provide a robust framework for dealing with \rev{spatially} 
correlated probabilities. We have been inspired by Tan et al.~\cite{tan2010decentralized} that used GRF to self-organize homogeneous robots. Besides the use of heterogeneous robots, there are other crucial differences between our work and~\cite{tan2010decentralized}: we model a continuous movement of the robots in a bounded environment and limit the robots' maximum velocities, while Tan et al. consider a discrete and a bounded environment and assume lattices as their environment. In addition, we introduce a different potential function that makes all the difference in our approach. In contrast with previous work, we adopt the Coulomb-Buckingham Potential~\cite{buckingham1938classical} coupled with a Kinetic Energy term to model the robots' interactions.

\section{Background}
\label{sec:background}
In this section, we overview some concepts about \textit{Gibbs Random Fields} (GRFs) explaining their properties and why they make sense in a swarm robotics context.
A GRF is a probabilistic graphical model that is a particular case of the  \textit{Markov Random Field} (MRF) when the joint probability density of the random variables is strictly positive. GRF models are based on local interactions between neighboring agents. The Markov property is a conditional property that allows one to ignore more distant information as soon as local information is provided. The Hammersley-Clifford theorem establishes the equivalence between \rev{a} MRF and a GRF~\cite{kindermann1980markov}. 

To describe these models succinctly, assume an undirected graph $\mathbf{G} = (\mathbf{V},\mathbf{E})$ with vertices as spatial sites and indexed by $v = 1, 2, . . . , \eta$. A random field on $\mathbf{G}$ is as collection of random variables $\mathbf{X} = \{X_{v}\}_{v \in \mathbf{V}}$ and, for each $v \in \mathbf{V}$, let $\Lambda_v$ be finite set called the phase space for site $v$ that represents where the random variable $X_v$ takes it values.
An instance of $\mathbf{X}$ establishes a state of the random field $x = \{(x_1,..., x_\eta): x_v \in \Lambda_v, v \in \mathbf{V}\}$ and the product space $\Lambda \triangleq \Lambda_1 \times ... \times \Lambda_\mathbf{\eta} $ forms the configuration space. 

A neighborhood system on $\mathbf{V}$ is a family $\mathcal{N} = \{\mathcal{N}_v\}_{v \in \mathbf{V}}$, where $\mathcal{N}_v \subset \mathbf{V}$ is the set of neighbors for site $v$ satisfying $v/\mathcal{N}_v$ and $r \in \mathcal{N}_v \Leftrightarrow v \in \mathcal{N}_r$. 
The neighborhood system induces the configuration of the undirected graph $\mathbf{G}$ by setting 
an edge $\{v,r\} \in \mathbf{E}$ between $v$ and $r$ if and only if $r \in \mathcal{N}_v$. A set $\mathcal{C} \subset \mathbf{V}$ is called a \textit{clique} if all elements of $\mathcal{C}$ are neighbors of each other. 

Thus, a random field $\mathbf{X}$ is called an MRF concerning the neighborhood system $\mathcal{N}$ if, $\forall v \in \mathbf{V}$,
\begin{equation}
\displaystyle{
    P(X_v=x_v \text{$|$} (X_r=x_r)_{r \neq v}) = P (X_v=x_v \text{$|$} (X_r=x_r)_{r \in \mathcal{N}_v})},
\label{eq:mrf}
\end{equation}
which indicates that the probability of the site $v$ assuming the state $x_v$ given the state of all other sites is equal to the probability of $v$ assuming the same state $x_v$ given only the states of neighboring sites. Such a definition reflects the local characteristics of the MRF constrained by the local Markov properties~\cite{koller2009probabilistic}\rev{. It is} convenient to model robotics swarms, since it implies the conditional independence of information coming from outside a neighborhood system, which supports the requirement of local interactions. 

\rev{A GRF is a particular application for (\ref{eq:mrf}) when the Gibbs measure can represent its joint probability density.} A Gibbs measure is a generalization of the canonical ensemble to infinite systems, which gives the probability of the system $\mathbf{X}$ being in the state $x$. Formally, let us denote a potential $\mathbf{U}$ as a family $\displaystyle{\{U_A : A \subset \mathbf{V}\}}$ of functions on the configuration space $\Lambda$, where $U_A : \Lambda \rightarrow \mathbb{R}$, and $U_A(x)$ depends only on $x_A \triangleq \{x_v: v \in A\}$. At the end, $U_A$ is only a function of the values at the sites contained in the set $A$, that is $\displaystyle{U_A(x) \equiv U_A(x_A)}$. In this way, given a potential $\mathbf{U}$, the potential energy $H(x)$ for configuration $x$ is defined as
\begin{equation}
H(x) = \sum \limits_{A \subset \mathbf{V}} U_A(x_A).
\label{eq:potentialenergy1}
\end{equation}

By definition~\cite{kindermann1980markov}, if $U_A \equiv 0$ whenever $A$ is not a \textit{clique} or a singleton, $\mathbf{U}$ is called a nearest-neighbor potential. If $U_A \equiv 0$ whenever $A$ is not a pair or a singleton, $\mathbf{U}$ is called a pairwise potential. $\mathbf{U}$ is called a pairwise, nearest-neighbor potential if it is both a pairwise potential and a nearest-neighbor potential. In particular, for a pairwise, nearest-neighbor potential $U$, we can write (\ref{eq:potentialenergy1}) as
\begin{equation}
H(x) = \sum \limits_{v \in \mathbf{V}} U_{\{v\}}(x_v) + \sum \limits_{(v,t) \in \mathbf{V}  \times \mathbf{V}, t \in \mathcal{N}_v} U_{\{v,t\}}(x_v, x_t).
\label{eq:potentialenergy2}
\end{equation}

Finally, a random field $\mathbf{X}$ is called a GRF if,
\begin{equation}
    P(\mathbf{X} = x) = \frac{1}{Z}e^{-\frac{H(x)}{T}}, \textrm{ with } Z = \sum \limits_{z} e^{-\frac{H(z)}{T}},
    \label{eq:gibbsdistribution}
\end{equation}
where $Z$ is the partition function (normalizing constant); $T$ is interpreted as temperature in the context of statistical physics; and $\frac{1}{Z}e^{-\frac{H(x)}{T}}$ is called Gibbs distribution. 

Researchers in statistical mechanics and mathematics usually applied the GRF to describe the distribution of system configurations at the thermodynamic equilibrium or measure the probability of such a system \rev{yielding} the desired state. One of the challenges of directly evaluating~(\ref{eq:gibbsdistribution}) is the high cardinality of the configuration space, which makes the computation of $Z$ intractable.

A typical approach to sequential sampling states in a configuration space given a probability function consists of using Markov Chain Monte Carlo (MCMC) methods, such as the Metropolis algorithm~\cite{hastings1970monte}. A process to \rev{parallelly} sample over~(\ref{eq:gibbsdistribution}) is described in the next section.

\section{Methodology}
\label{sec:methodology}
The general idea of our methodology consists of modeling the configuration of a swarm of heterogeneous robots as a GRF and then sampling velocities for each robot in a decentralized way, which leads the entire swarm to a convergence towards the global minimum of the potential.

\subsection{Formalization}
Consider a set $\mathcal{R}$ of $\eta$ heterogeneous robots navigating in a bounded region within the two-dimensional Euclidean space\footnote{We assume two-dimensional space for convenience but one can straightforward extend it to three-dimensional space.}. \rev{The state of the $i$-th robot} 
at time step $t$ is represented by its pose $\mathbf{q}_i ^{(t)}$ and velocity\footnote{From now on, we use the symbol $\mathbf{v}$ to represent robot velocities.} $\dot{\mathbf{q}_i}^{(t)} = \mathbf{v}_i^{(t)}$, which is bounded by $v_{max}$, $||\mathbf{v}_i^{(t)}|| \leq v_{max}$. In addition, robots are driven by a holonomic kinematic model with motion model $\displaystyle{\mathcal{K} : (\mathbf{q}_i^{(t)}, \mathbf{v}_i^{(t)}) \rightarrow (\mathbf{q}_i^{(t+1)})}$. The heterogeneity of the system is modeled by a partition $\tau = \{{\tau}_1, ..., {\tau}_m \}$, with each ${\tau}_k \subset \mathcal{R}$ containing exclusively all robots of type $k$. That is, $\forall (j,k) : j \neq k \ \rightarrow {\tau}_k \cap {\tau}_j = \emptyset$. 

Each robot has a circular sensing range of radius $\lambda$, where it can estimate the relative position and velocity of other robots as well as their type, and also obstacles within the environment. 
Obstacles are represented as a finite set of points $\displaystyle{\mathcal{O} = \{\mathbf{o}_1, ..., \mathbf{o}_n\}}$. An obstacle detected by the \rev{$i$-th} robot consists of a subset of points $\mathcal{O}_i \subset \mathcal{O}$, where $\displaystyle{ \mathbf{o}_j \in \mathcal{O}_i \rightarrow ||\mathbf{o}_j - \mathbf{q}_i|| \leq \lambda}$ and $||\mathbf{o}_j - \mathbf{q}_i||$ is the Euclidean norm between two points. 

The neighborhood system for the $i$-th robot, constrained by the sensing range $\lambda$, defines a set of robots: 
\begin{equation}
    \mathcal{N}_i \triangleq \{j \in \mathcal{R}: j \neq i, ||\mathbf{q}_j - \mathbf{q}_i|| \leq \lambda \}.
    \label{eq:neighborhood}
\end{equation}

\subsection{Extension of the GRF to swarm robotics}
Inspired by the GRF capability in modeling local interactions, here we discuss its concepts in the context of swarm robotics. 
Following the modeling presented in section~\ref{sec:background}, let us define a graph $\mathbf{G} = (\mathcal{R},\mathbf{E})$ with a set of random variables $\mathbf{X} = \{X_i\}_{i \in \mathcal{R}}$, in which each $X_i$ models the random velocity $\mathbf{v}_i$ of \rev{the $i$-th robot}. A configuration of the system $\mathbf{X}$ is $\displaystyle{\mathbf{x} = \{\mathbf{v}_1,... ,\mathbf{v}_\eta\}}$ where $\mathbf{v}_i \in \Lambda_i$ and represents the velocities performed by each robot.

A neighborhood system on $\mathcal{R}$, given a configuration space $\mathbf{x}$, is a family $\displaystyle{\mathcal{N} = \{\mathcal{N}_{i}\}_{i \in \mathcal{R}}}$, where $\mathcal{N}_i \subset \mathcal{R}$ is the set of neighbors defined in (\ref{eq:neighborhood}) and satisfies $i/\mathcal{N}_i$ and the symmetry $\displaystyle{j \in \mathcal{N}_i \Leftrightarrow i \in \mathcal{N}_j}$. The neighborhood system $\mathcal{N}$ induces the configuration of the graph $\mathbf{G}$ by establishing an edge between each pair $(i,j)$ of robots if and only if $j \in \mathcal{N}_i$.

Until now, our definitions only allow us to calculate the probability of the entire swarm reaching a certain configuration, but what we require here is to sample velocities for each robot given the information about the robots in its neighborhood. Next, we explain how we perform such \rev{a} procedure in a decentralized way using the Gibbs distribution.

\subsection{Parallel Gibbs sampling}
Parallel Gibbs sampling implies that all robots are simultaneously updating their velocities based on the configuration $\mathbf{x}$ at time $t$. Such \rev{a} method is possible here due to the local nature of the Gibbs potential energy. 

Formally, let $t$ denote the temporal index and $\displaystyle{\mathbf{x}^{(t)} = \mathbf{x} = (\mathbf{v}_1, ..., \mathbf{v}_\eta)}$ be the swarm configuration at time $t$. Let $\mathbf{Z}_i(\mathbf{x}) \triangleq \{\mathbf{z}_i : ||\mathbf{z}_i|| \leq v_{max}\}$, where $\mathbf{Z}_i(\mathbf{x}) \subset \Lambda_i$, be the set of possible velocities for \rev{the $i$-th robot} given the configuration $\mathbf{x}^{(t)}$.
Using~(\ref{eq:gibbsdistribution}), the $i$-th robot updates its velocity $\mathbf{v_i}^{(t)} = \mathbf{v_i}$ to $\mathbf{v_i}^{(t+1)} = \mathbf{\bar v_i}$ with probability
\begin{equation}
    P_i(\mathbf{v_i}, \mathbf{\bar v_i}|\mathbf{x}) = 
    \begin{cases} 
    \frac{e^{-H(\mathbf{\bar v_i}, \mathbf{x}_{\mathcal{R} \setminus i})/T}}{\sum \limits_{\mathbf{z}_j \in \mathbf{Z}_i(x)} e^{-H(\mathbf{z}_j, \mathbf{x}_{\mathcal{R} \setminus i})/T}},& \text{ if } \mathbf{\bar v_i} \in Z_i(\mathbf{x}) \\
    0,& \text{otherwise}.
    \end{cases}
    \label{eq:probglobal}
\end{equation}
Note that (\ref{eq:probglobal}) still depends on the global knowledge at the potential energy $H(\cdot , \mathbf{x}_{\mathcal{R} \setminus i})$. However, if we rewrite~(\ref{eq:potentialenergy2}) as,
\begin{equation}
    \begin{split}
        H(\cdot , \mathbf{x}_{\mathcal{R} \setminus i}) = 
            \left( U_{\{i\}}(\cdot) + \sum \limits_{\forall j \in \mathcal{R} \setminus i} U_{\{j\}}(\mathbf{v}_j) \right) + \\
            \sum \limits_{\forall j \in \mathcal{N}_i}U_{\{i,j\}}(\cdot, \mathbf{v_j}),
    \end{split}
\end{equation}
one may note that the second term inside the parenthesis is constant for \rev{the $i$-th robot}, which lets us reduce the form for $H(\mathbf{\bar v_i} , \mathbf{x}_{\mathcal{R} \setminus i})$ and $H(\mathbf{z_i} , \mathbf{x}_{\mathcal{R} \setminus i})$ in~(\ref{eq:probglobal}). This shows the local nature of the Gibbs potential energy and implies that we do not require the knowledge of the entire swarm to sample velocities for \rev{the $i$-th robot}, but only information about its neighbors $\mathcal{N}_i$. Thus, we can rewrite~(\ref{eq:probglobal}) as 
\begin{equation}
    P_i(\mathbf{v_i}, \mathbf{\bar v_i}|\mathbf{x}) = 
    \frac{e^{-\left(U_{\{i\}}(\mathbf{\bar v_i}) + \sum \limits_{\forall j \in \mathcal{N}_i}U_{\{i,j\}}(\mathbf{\bar v_i}, \mathbf{v_j})\right)T^{-1}}}
    {\sum \limits_{\mathbf{z_i} \in \mathbf{Z}_i(x)} e^{-\left(U_{\{i\}}(\mathbf{z_i}) + \sum \limits_{\forall j \in \mathcal{N}_i}U_{\{i,j\}}(\mathbf{z_i}, \mathbf{v_j})\right)T^{-1}}},
    \label{eq:problocal}
\end{equation}
where $\mathbf{\bar v_i} \in \mathbf{Z}_i(\mathbf{x})$.

\subsection{Potential energy}
Here we propose \rev{combining} 
two potential functions into the potential energy $H(\cdot)$ to achieve simultaneous segregative-flocking behaviors of the swarm.
\subsubsection{Coulomb-Buckingham potential}
The Coulomb-Buckingham potential~\cite{buckingham1938classical} is a combination of the Lennard-Jones potential with the Coulomb potential used to describe the interaction among particles considering their charges. We took advantage of such a mechanism to model the \rev{swarm's heterogeneity} 
by setting the particle charges. The formula for the interaction is
\begin{equation}
\Phi (r)=\varepsilon \left({\frac {6}{\alpha -6}}e^{\alpha} \left(1-{\frac {r}{r_{0}}}\right)-{\frac {\alpha }{\alpha -6}}\left({\frac {r_{0}}{r}}\right)^{6}\right) + \frac {c_{i}c_{j}}{4\pi \varepsilon_0 r},
    \label{eq:cbpotential}
\end{equation}
where $r = ||\mathbf{q}_j - \mathbf{q}_i||$ is the euclidean distance between the particles $i$ and $j$; $\varepsilon$ is the depth of the minimum energy; $r_{0}$ is the minimum energy distance; $\alpha$ is a free dimensionless parameter; $c_i$ and $c_j$ are the charges of the particles $i$ and $j$; and $\varepsilon_0$ is an electric constant. 

\begin{figure}[t]
		\centering
		\includegraphics[width=.75\columnwidth]{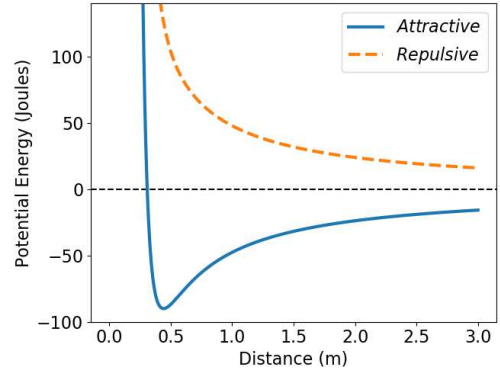}
	\caption{The Coulomb-Buckingham potential function. It depends on the distance $r$ among the $i$-th and $j$-th robots and a function $C(i,j)$ that produces attractive or repulsive behaviors depending on the heterogeneity among them. 
	} 
	\label{fig:cbpotential}
\end{figure}

We define the interaction among \rev{the $i$-th and $j$-th robots} by replacing the product $c_i c_j$ by the following function,
\begin{equation}
    C(i,j) = \left(2~\mathbb{1}((i, j) \in \tau_k ) - 1 \right)|c_i c_j|,
\end{equation}
where $\mathbb{1}(\cdot)$ denotes the indicator function. In this way $C(i,j)$ will be positive if the $i$-th and $j$-th robots belong to the same group $\tau_k$ and negative otherwise. Fig.~\ref{fig:cbpotential} illustrates the Coulomb-Buckingham potential.

\subsubsection{Kinetic energy}
We assume classical mechanics to compute the kinetic energy produced by relative velocities of the $i$-th robot neighbors. Let $\mathbf{V}_i$ define the resultant of the relative velocities among all neighbors within the same partition of \rev{the $i$-th robot}. The kinetic energy $\mathbf{E_k}$ relative to the $i$-th robot is:
\begin{equation}
\mathbf{V}_i = \sum \limits_{\forall j \in \mathcal{N}_i \land (i,j) \in \tau_k} \mathbf{v}_j, \ \ \ \mathbf{E_k}(\mathbf{V}_i) = \frac{1}{2} m (\mathbf{V}_i \cdot \mathbf{V}_i),
\end{equation}
where $m$ is the cumulative mass of the group.

\subsubsection{Combination}
 We combine the Coulomb-Buckingham potential and the kinetic energy to define the potential energy $H(\mathbf{\bar v}_i , \mathbf{x}_{\mathcal{R} \setminus i})$. Here, the individual potential $U_{\{i\}}(\mathbf{\bar v_i})$ represents an obstacle avoidance factor defined by
\begin{equation}
    \begin{split}
            U_{\{i\}}(\mathbf{\bar v_i}) = \sum \limits_{\forall j \in \mathcal{O}_i} \Phi(||\mathcal{K}(\mathbf{q}_i, \bar{\mathbf{v}_i}) - \mathbf{o_j}||),
    \end{split}
    \label{eq:paiwisepot}
\end{equation}
and $\forall j \in \mathcal{O}_i: C(i, j) > 0$.

The nearest-neighbor potential establishes a segregative-flocking factor formulated as
\begin{equation}
    \begin{split}
            \sum \limits_{\forall j \in \mathcal{N}_i}U_{\{i,j\}}(\mathbf{\bar v_i}, \mathbf{v_j}) = 
            \sum \limits_{\forall j \in \mathcal{N}_i} \Phi(||\mathcal{K}(\mathbf{q}_i, \bar{\mathbf{v}_i}) - \mathcal{K}(\mathbf{q}_j, {\mathbf{v}_j})||) + \\
            \mathbf{E_k}(\mathbf{V}_i) + \mathbf{E_k}(v_{max} - \bar{\mathbf{v}_i}),
    \end{split}
    \label{eq:nearestneighborpot}
\end{equation}
where the first term defines the attraction and repulsion among two robots and the second one is the relative velocity of the neighbors. Since we compute the kinetic energy using relative velocities, when $\mathbf{E_k}(\mathbf{V}_i) \rightarrow 0$, there is a duality on the behavior produced by the robots. More specifically, one may not differentiate if the robots are stationary or moving with the same velocities. To avoid such duality, we added a third term to force \rev{the $i$-th robot} to reach its maximum speed.

\subsection{Sampling algorithm}
Finally, given the probability function~(\ref{eq:problocal}) and the potential energy defined by combination of (\ref{eq:paiwisepot}) and~(\ref{eq:nearestneighborpot}) one may use a MCMC algorithm to sample velocities for \rev{the $i$-th robot}. In this work, we use the Metropolis-Hastings algorithm~\cite{peskun1973optimum} for sampling the velocities.

%
\begin{figure*}[t]
	\begin{subfigure}{.178\textwidth}
		\centering
		\includegraphics[width=.96\linewidth]{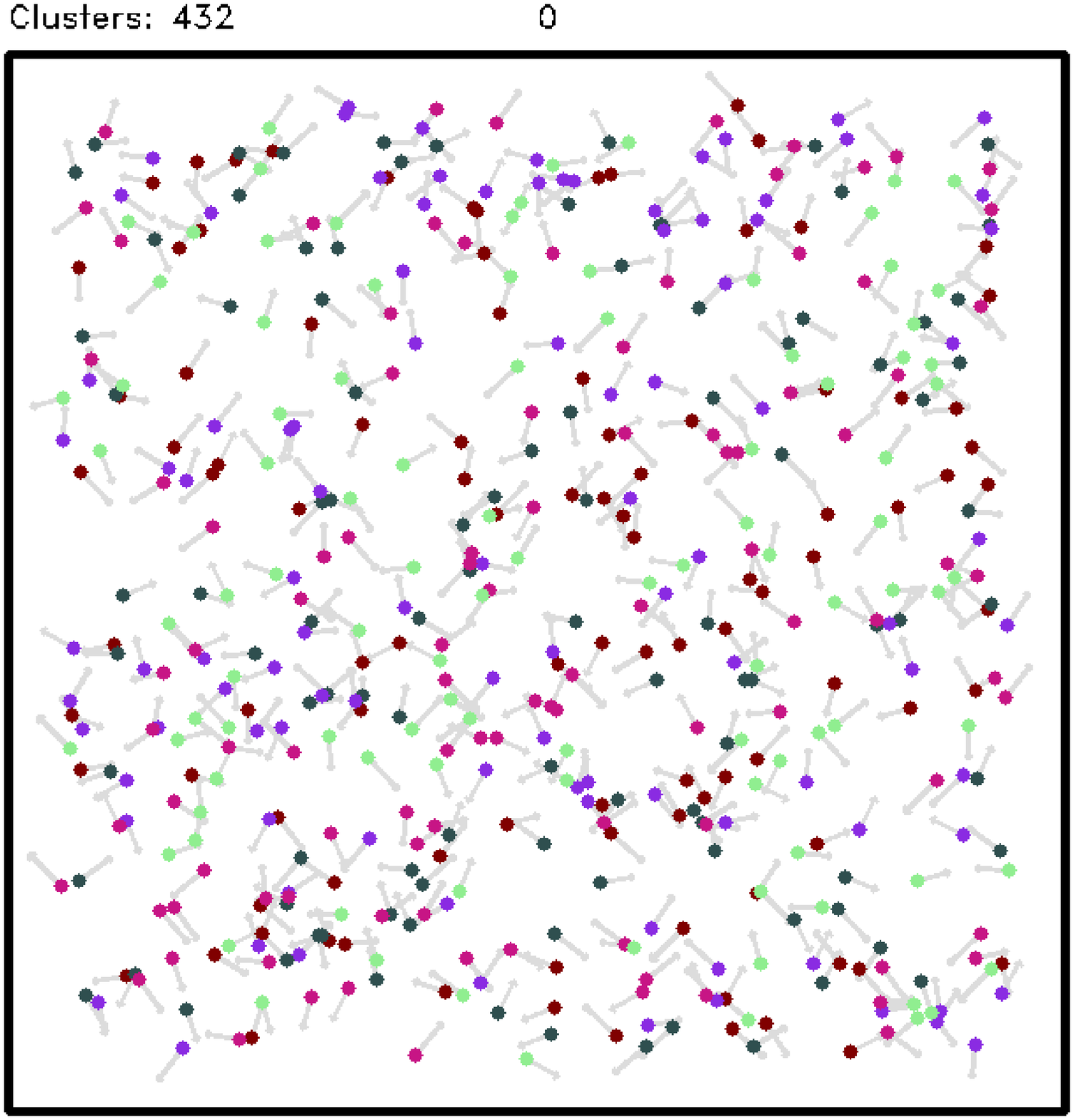}
		\caption{$n=0$}
	\end{subfigure}%
	\hspace*{-1.2em}
	\begin{subfigure}{.178\textwidth}
		\centering
		\includegraphics[width=.96\linewidth]{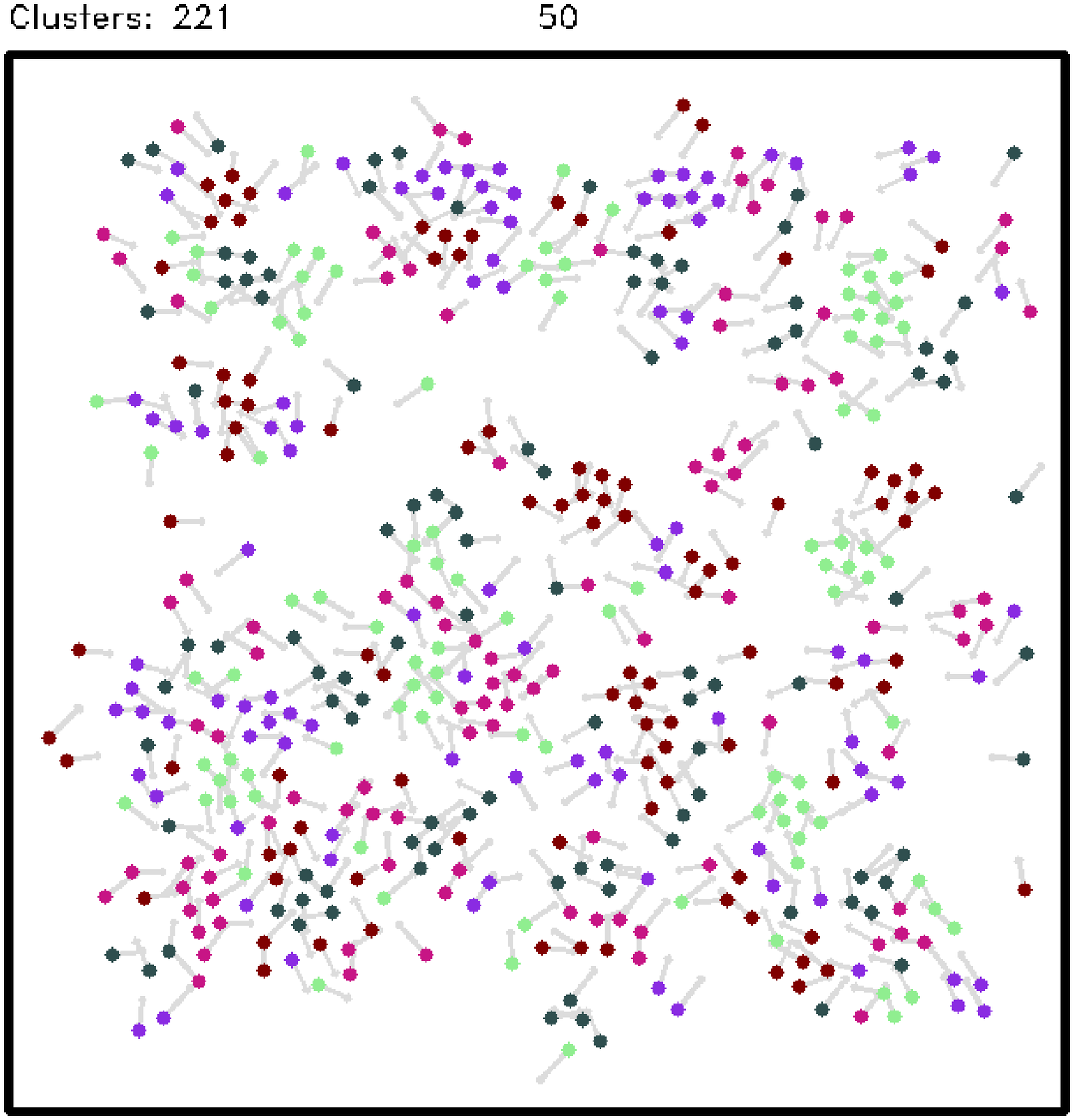}
		\caption{$n=50$}
	\end{subfigure}
	\hspace*{-1.6em}
	\begin{subfigure}{.178\textwidth}
		\centering
		\includegraphics[width=.96\linewidth]{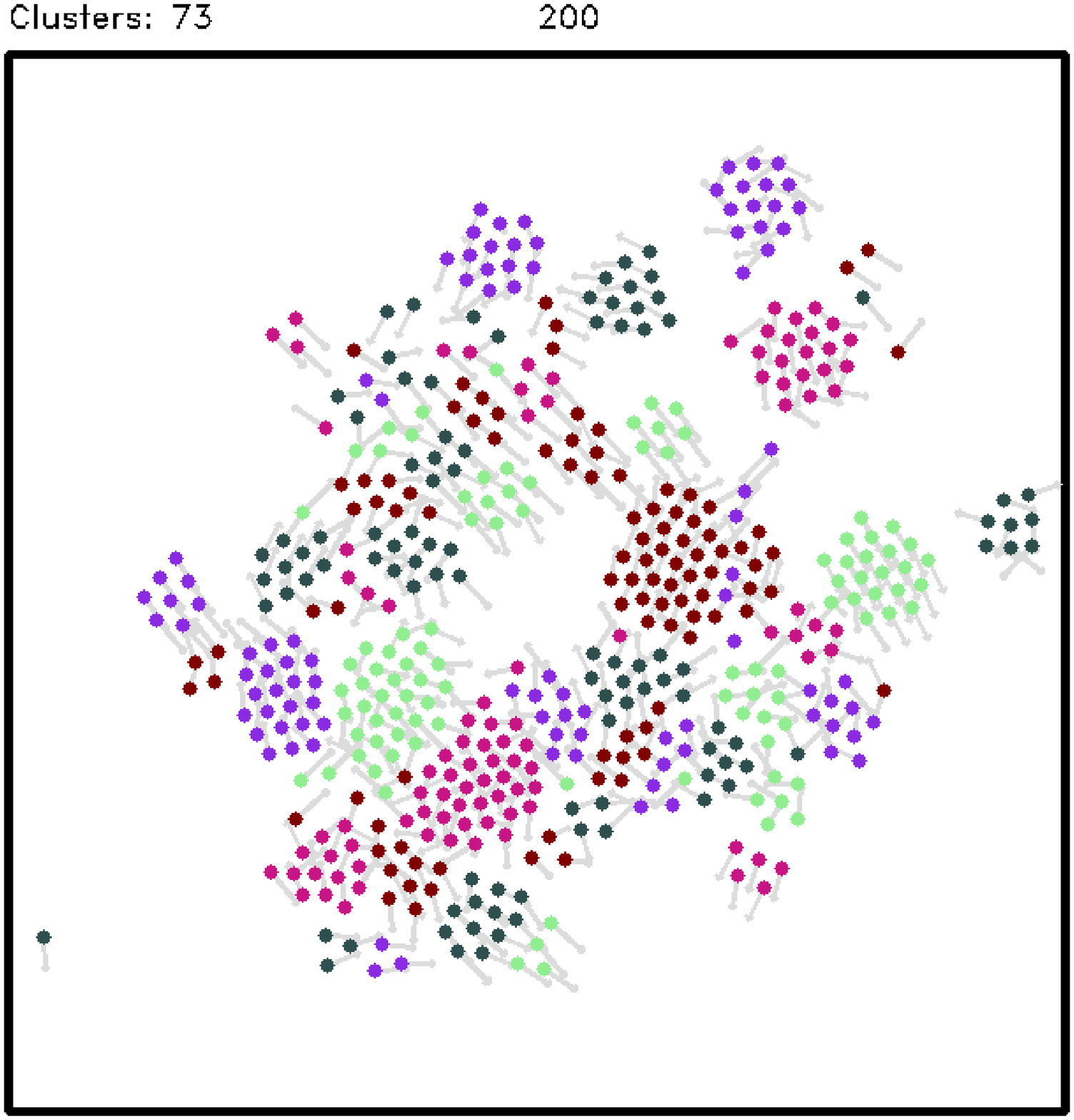}
		\caption{$n=200$}
	\end{subfigure}
	\hspace*{-1.6em}
	\begin{subfigure}{.178\textwidth}
		\centering
		\includegraphics[width=.96\linewidth]{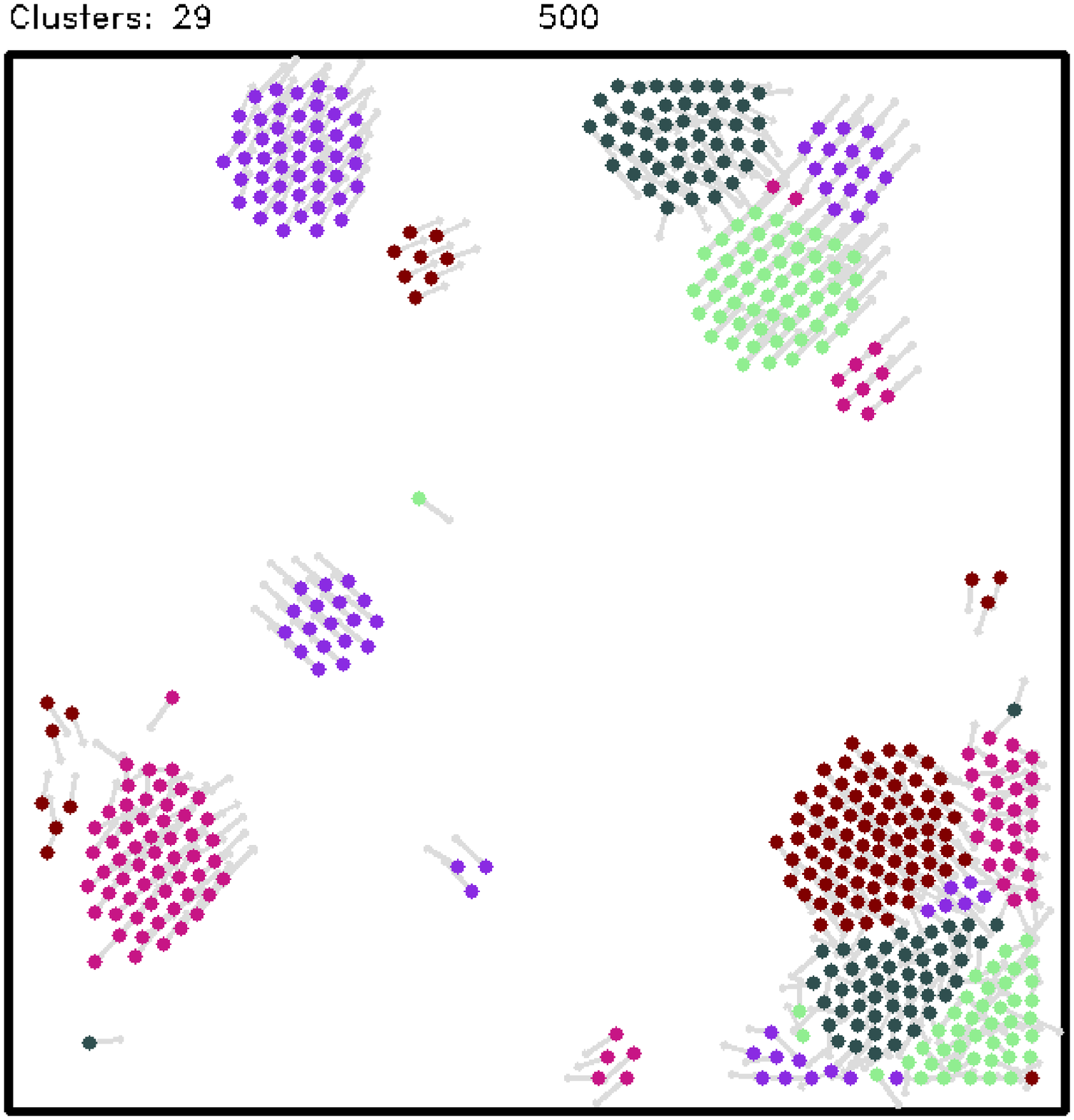}
		\caption{$n=500$}
	\end{subfigure}
	\hspace*{-1.6em}
	\begin{subfigure}{.178\textwidth}
		\centering
		\includegraphics[width=.96\linewidth]{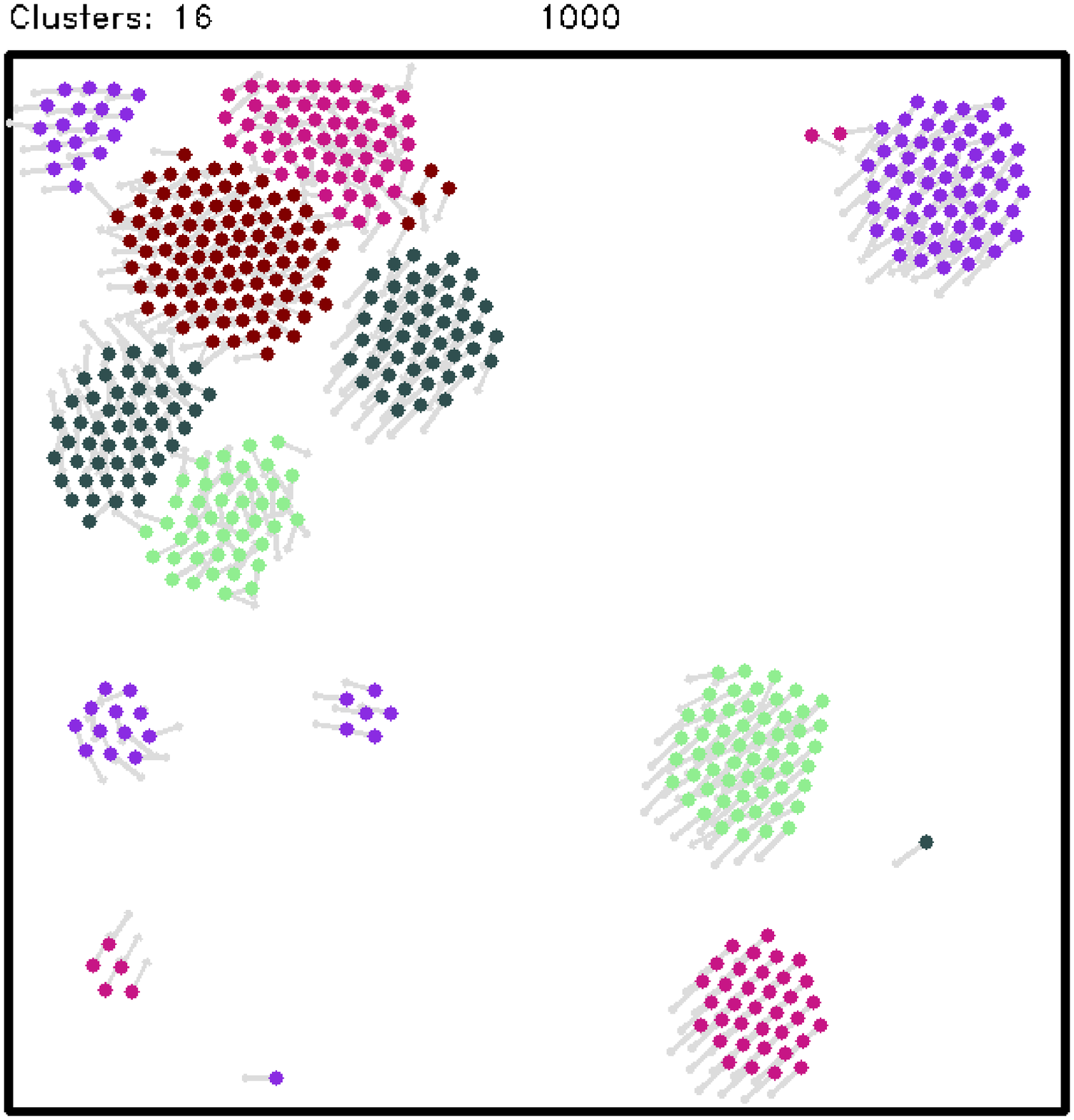}
		\caption{$n=1000$}
	\end{subfigure}
	\hspace*{-1.6em}
	\begin{subfigure}{.178\textwidth}
		\centering
		\includegraphics[width=.96\linewidth]{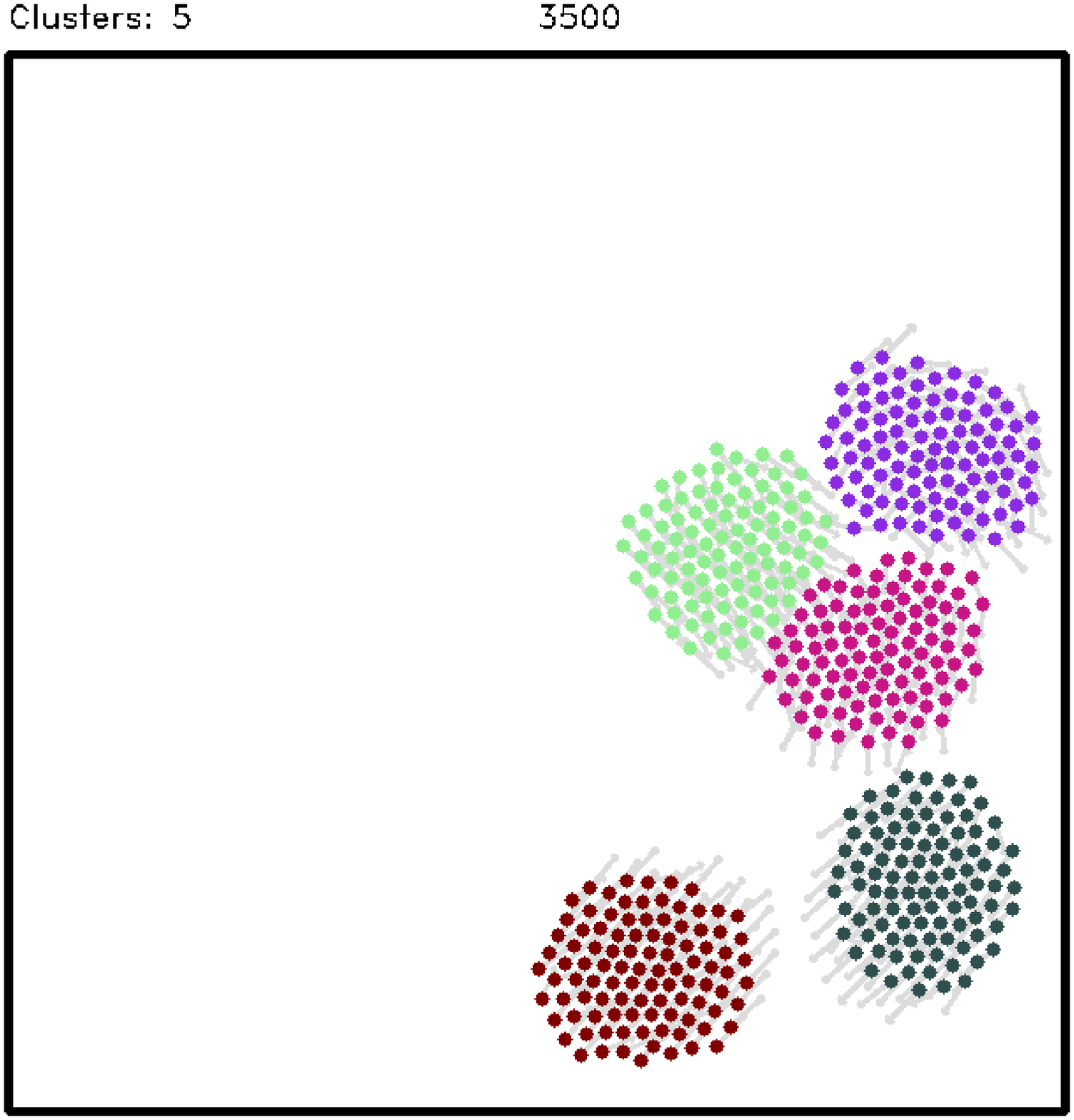}
		\caption{$n=3500$}
	\end{subfigure}
	\caption{Demonstration of segregative flocking of $5$ heterogeneous groups with $100$ robots each. }
	\label{fig:snapshots}
\end{figure*}

\section{Experiments and Results}
\label{sec:experiments}
To evaluate the performance of our approach, we conduct a series of simulated experiments. We first analyze the segregative behavior by measuring the \rev{method's performance} 
for different configurations and comparing the results with other methods from the literature. Then, to analyze the flocking capabilities, we evaluate the velocity consensus \rev{and} 
the cohesion among the robots when there is noise in the sensor. Finally, we performed experiments with real robots as a proof-of-concept to show the feasibility of our approach in real scenarios. 
Fig.~\ref{fig:snapshots} shows snapshots illustrating the simultaneous flocking segregative behavior produced by our methodology. A video of the experiments is available at Youtube\footnote{\url{https://youtu.be/KooNGIStWlM}} and the source code at Github\footnote{\url{https://github.com/verlab/2021-icra-grf-swarm}}.

\subsection{Segregation Analysis}
To evaluate the segregative behavior considering only local information, we compare the convergence rate of our approach against the one presented by Mitrano et al.~\cite{mitrano2019minimalistic} and Inácio et al.~\cite{inacio2019pso}. We consider the work proposed by Santos et al.~\cite{santos2020spatial} as a baseline since it assumes global knowledge about the positions of other robots leading to a fast convergence rate. We also contrast our methodology with a deterministic gradient descent approach using potential differentials to show that such mechanism may be easily trapped at local minima.

\begin{figure}[h]
	\begin{subfigure}{.465\textwidth}
		\centering
		\includegraphics[width=.95\linewidth]{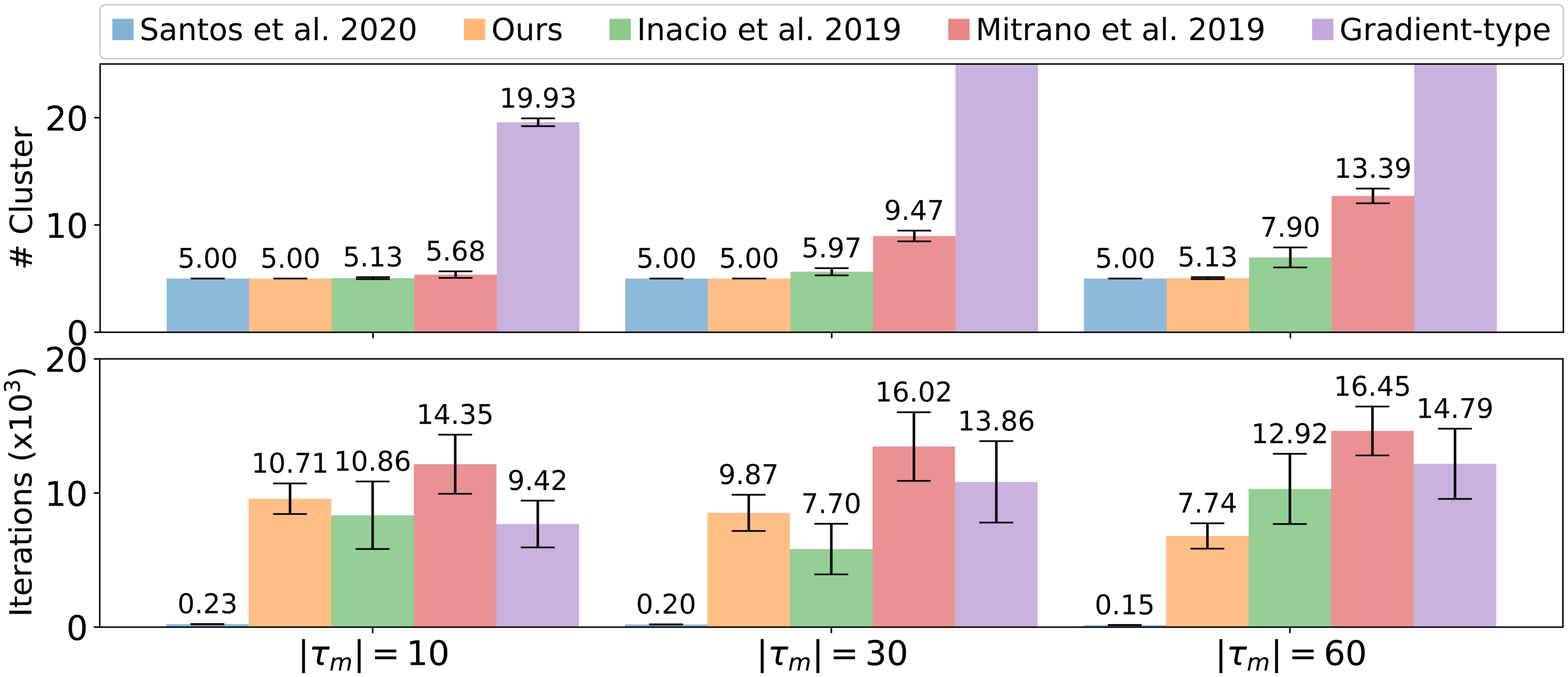}
		\caption{}
		\label{fig:rangerobots}
	\end{subfigure}%
	\qquad
	\begin{subfigure}{.465\textwidth}
		\centering
		\includegraphics[width=.95\linewidth]{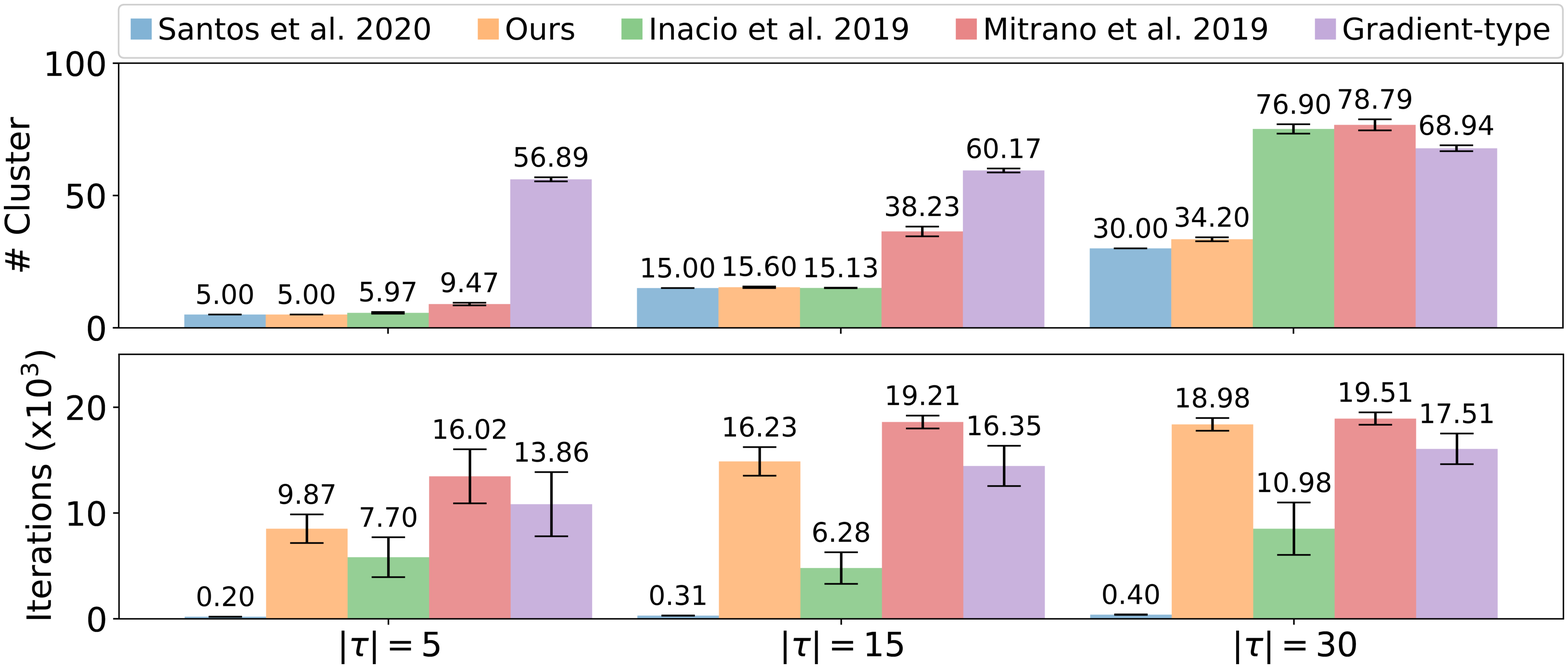}
		\caption{}
		\label{fig:rangegroups}
	\end{subfigure}%
	\caption{The minimum number of clusters yield by each approach in up to $20000$ iterations when: (a) we increase the number of robots $|\tau_m| = \{10,30,60\}$ keeping $|\tau|=5$ heterogeneous groups; and (b) we increase the number of groups $|\tau|=\{5, 15, 30\}$ keeping $|\tau_m| = 10$ robots per group.
	} 
	\label{fig:segregationexp}
\end{figure}

The experiments consisted of $100$ runs of each approach with a maximum of $20000$ iterations. \rev{A random initial state is generated for each run, but it is the same for all approaches.} 
At each iteration, the robot \rev{can} move a maximum of $0.02$ meters in a square area of $10$ by $10$ meters with the walls being considered obstacles. We varied the number of robots and the number of heterogeneous groups
to evaluate each \rev{approach's performance.} 
As a metric, we compute the total amount of clusters formed by robots of the same type \rev{and} 
the number of iterations necessary to reach it. Here, two robots of the same type are considered to be in the same cluster if their relative distance is less than $0.3$ meters -- the robot radius is $0.07$ meters. The sensing range is set to $0.5$ meters ($\lambda=0.5$) Fig.~\ref{fig:segregationexp} shows the mean and the $99\%$ confidence interval comparing one with the other approaches.

Analyzing the segregation using the number of formed clusters, we can see that all approaches executed relatively well for a small number of groups, even for an increasing number of robots per group (Fig. \ref{fig:rangerobots}, top). \rev{The exception is the Gradient-Descent method, which gets trapped in local minima and cannot reach a segregated state.} 
When the number of groups increases, our approach significantly outperforms the others, with a performance close to the baseline which uses global information (Fig. \ref{fig:rangegroups}, top). When there is a large number of groups, robots usually get trapped by other groups and cannot reach a segregated state. By relying on the stochastic nature of the GRF, our approach \rev{can handle these situations better.} 

Regarding the performance in terms of \rev{the} number of iterations to reach segregation, we can see that the methods have a similar performance \rev{on} average when increasing the number of robots. \rev{However, all} of them are significantly slower than the baseline, which uses global information (Fig. \ref{fig:rangerobots}, bottom). When we vary the number of groups, we can see that Inácio et al. \rev{have} a better performance (Fig. \ref{fig:rangegroups}, bottom). However there may be a caveat, specially for $|\tau| = 30$: our metric considers the number of iterations spent until reaching the minimum number of clusters. As previously mentioned, Inácio's method does not reach the minimum number of clusters \rev{on} several occasions. So, it may be converging faster but to a sub-optimal configuration. \rev{On the other hand, our method may take} 
longer due to its stochastic nature, but has a much better success rate. 

\subsection{Flocking Analysis}
\begin{figure}[h]
	\centering
	\includegraphics[width=.87\linewidth]{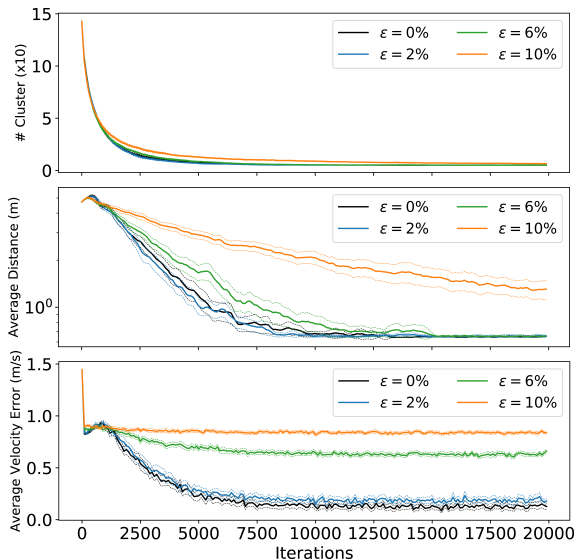}
	\caption{Impact of the noise in the sensor model \rev{on} the performance of our method. The graphics display the number of clusters yield, the average distance, and velocity error among \rev{the same group of robots} in up to $20000$ iterations.}
	\label{fig:navigation}
\end{figure}

To evaluate the effectiveness of our approach in producing flocking behaviors, we carried out some experiments and \rev{analyzed} them regarding the average distance (cohesion) and consensus speed between robots of the same type. The robustness of our method is assessed by adding Gaussian noise $\epsilon$ to the sensor model so that the relative position and velocity estimates are not reliable. Here, we perform $100$ runs with a maximum of $20000$ iterations. We assume $|\tau_m| = 30$ robots into $|\tau|=5$ heterogeneous groups and sensing range $\lambda=0.5$ meters. The robots start each run in a random initial state and perform both segregation and flocking within an environment of 10 by 10 meters at a maximum speed of $v_{max}=1.0$ meters per second. Noise in the sensor model ranges from $\epsilon = \{0\%, 2\%, 6\%, 10\%\}$ for both relative position and velocity. A noise of $\epsilon = 10\%$ implies an error of up to $10\%\lambda = 0.05$ meters in position and $10\%v_{max} = 0.10$ meters per seconds in speed. Fig.~\ref{fig:navigation} shows the mean and the $95\%$ confidence interval evaluating the impact that such a noise causes in our methodology.

As expected, increasing noise in the sensor model significantly impacts the velocity consensus. We observed that up to $\epsilon = 6\%$, the swarm is able to maintain the flocking-segregative behavior for the \rev{experiments' configuration}. 
When $\epsilon = 10\%$, we notice the velocity consensus degrading and, consequently, the flocking behavior does not converge. However, even with \rev{the} noise we \rev{can} 
segregate the swarm to the minimal number of clusters most of the times for a sufficiently large number of iterations.

\subsection{Real robots}
To evaluate the feasibility of our approach in a real environment, we performed proof-of-concept experiments using five e-puck robots~\cite{mondada2009puck}. The robots receive velocity commands from a remote server executing ROS. Given that our robots do not have any sensor that allows them to estimate the relative position and velocity of neighboring robots, we emulate such a sensor using the Optitrack motion capture system~\cite{optitrack}.

\begin{figure}[h]
	\centering
	\includegraphics[width=.87\linewidth]{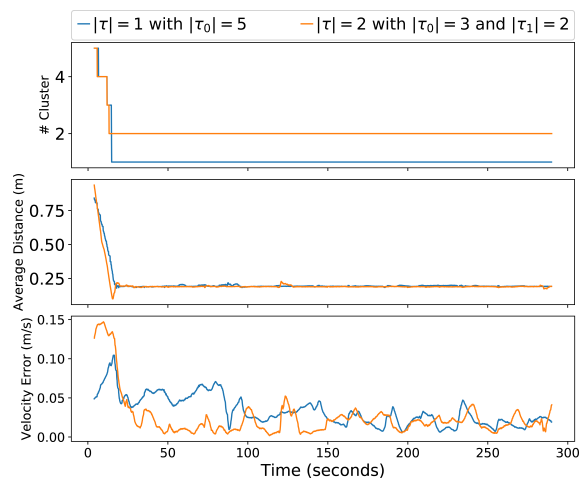}
	\caption{Results on real experiments using five epuck robots divided into one or two groups. }
	\label{fig:realplot}
\end{figure}

Here we consider the bounded environment as a square area of 2 by 2 meters restricted by walls. We set the sensing distance to $\lambda = 0.3$, and as we have only a few robots, we evaluate cases where we have one or two groups. That is one group with five robots and two groups with two and three robots each. Fig.~\ref{fig:realplot} shows the performance of our approach using real robots. We can see that the robots can reach segregation and also keep the flocking behavior, but with some noise in the velocity consensus due to the uncertainties observed in real settings. 

\section{Conclusion and Future Work}
\label{sec:conclusion}
\rev{This paper presented a novel decentralized approach that allows a swarm of heterogeneous robots to achieve simultaneously segregation and flocking behaviors using only local sensing.}
%
We compared the 
segregative behavior with some \rev{state-of-the-art approaches and evaluated the flocking behavior in simulated and real scenarios.}
%
Results showed that our methodology \rev{can 
segregate} a group of heterogeneous robots while keeping \rev{cohesive} 
navigation around the environment.

\rev{In future work, we intend to equip our robots with distance and bearing sensors and temporally combine their information to locally estimate the neighboring robots' velocity and position.}
Moreover, there are several opportunities for future studies and applications using the GRF framework. In particular, we intend to investigate the possibility of performing more complex tasks, such as transport and shape-formation. 

%



\clearpage
\bibliographystyle{unsrt}
\bibliography{root}

\end{document}